\documentclass[10pt,conference]{IEEEtran}
\makeatletter

\IEEEoverridecommandlockouts                              

\usepackage{amssymb, amsmath}
\usepackage{graphicx}
\usepackage[vlined,ruled]{algorithm2e}
\usepackage{color}
\usepackage{multirow}
\usepackage{float}
\usepackage{verbatim}
\usepackage{tabularx}
\usepackage{url}

\usepackage{subfig}
\usepackage{xspace}
\usepackage{balance}
\usepackage{makecell}
\usepackage{tabularx}

\DeclareMathOperator*{\argmax}{argmax}
\newcommand{\system}{TCAR\xspace}

\begin{document}

\title{Your instruction may be crisp, but not clear to me!}

\author{\IEEEauthorblockN{Pradip Pramanick, Chayan Sarkar, and Indrajit Bhattacharya}
	\IEEEauthorblockA{TCS Research \& Innovation, India}
}

\maketitle
\thispagestyle{empty}
\pagestyle{empty}

\begin{abstract}
The number of robots deployed in our daily surroundings is ever-increasing. Even in the industrial set-up, the use of coworker robots is increasing rapidly. These cohabitant robots perform various tasks as instructed by co-located human beings. Thus, a natural interaction mechanism plays a big role in the usability and acceptability of the robot, especially by a non-expert user. The recent development in natural language processing (NLP) has paved the way for chatbots to generate an automatic response for users' query. A robot can be equipped with such a dialogue system. However, the goal of human-robot interaction is not focused on generating a response to queries, but it often involves performing some tasks in the physical world. Thus, a system is required that can detect user intended task from the natural instruction along with the set of pre- and post-conditions. In this work, we develop a dialogue engine for a robot that can classify and map a task instruction to the robot's capability. If there is some ambiguity in the instructions or some required information is missing, which is often the case in natural conversation, it asks an appropriate question(s) to resolve it. The goal is to generate minimal and pin-pointed queries for the user to resolve an ambiguity. We evaluate our system for a telepresence scenario where a remote user instructs the robot for various tasks. Our study based on 12 individuals shows that the proposed dialogue strategy can help a novice user to effectively interact with a robot, leading to satisfactory user experience.
\end{abstract}

\section{INTRODUCTION}

Factories are using various robots as part of their workforce for decades. The use is mainly restricted to a specific area for predefined, repetitive jobs. Recently, we see a large number of coworker robots are deployed in industrial setup along with robots in our daily surrounding like home, office, restaurant, airport, shopping centers, etc~\cite{el2018working,pramanick2018defatigue}. Often these cohabitant robots have to interact with human beings. Thus, a natural conversation mechanism is a necessity for these robots for better usability and acceptability by the users. 

In recent times, the deployment of a chatbot by various businesses and organizations has increased rapidly. They are usually trained with a vast amount of domain knowledge and they perform query answering from this structured knowledge. They are equipped with customized natural language processing (NLP) tool-sets that help to extract the input data from a conversation. A robot deployed in our surrounding can utilize such a chatbot to derive the human intention and set a goal for itself. As the functionality of a robot is not limited for question-answering only, but to perform a certain type of tasks within its capability, the goal setting for self often involves action planning. If a robot accepts tasks through natural instructions, it needs to identify the intended task and generate a plan to perform the task. Unlike constraint factory floors, where the robot performs a predefined sequence of tasks, run-time task identification and planning is required. 

\begin{figure}
	\centering
	\includegraphics[width=\linewidth]{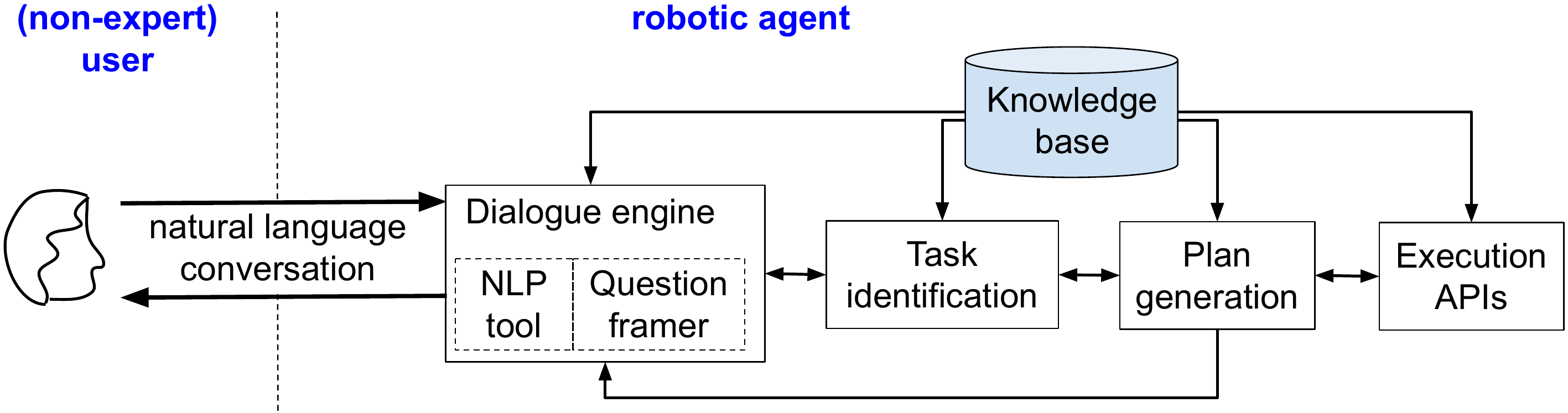}
	\caption{Overview of robotic agent for task identification and execution through natural language dialogue.}
	\label{fig:dialogue-overview}
\end{figure}

In this work, we develop a dialogue engine for robots and an overview is shown in Fig.~\ref{fig:dialogue-overview}. The dialogue engine in its full capacity would be able to capture human intended tasks through audio command and gesture, engage in dialogue (only when it is necessary) if there is ambiguity/missing information in natural interaction, and generate an executable plan to complete the task.
Scope of this work is limited to text-based (natural language) input-output. However, any audio-to-text and text-to-audio system can be coupled with this for vocal conversation. 

Most of the existing chatbots are trained with query-response pairs and a given query is classified to such a predefined pair. A robotic task instruction requires a set of pre- and post-conditions to be satisfied, which varies significantly with the number of conditions and task context. As a result, the predominant approach of classifying a task instruction to a predefined task-action pair is not sufficient. Also, the ambiguity present in natural conversation cannot be handled using a predefined query. Thus, we develop a dialogue strategy to generate a context-specific query, should there be any ambiguity and/or missing information in the conversation. Our dialogue strategy allows a human to control the dialogue flow by specifying such an intention. Since training data for robotic tasks instructions are scarce for most application domains, we use a set of probabilistic classifiers that does not require a large volume of training data. 

Our major contributions are two-fold. Firstly, we develop a mixed-initiative dialogue engine for a robot that can identify tasks along with all the parameters from a natural conversation and generate a viable plan to execute it. Secondly, we develop a dialogue strategy that resolves ambiguity and failure in task understanding with a minimal query.

\section{RELATED WORK}
\label{sec:related}
Advancements in deep learning and reinforcement learning have empowered many complex conversational systems~\cite{chen2017survey}, for both domain-specific, task-oriented dialogue~\cite{patidar2018automatic} and general-purpose dialogue for social conversation~\cite{yan2018chitty}. Social dialogue agents essentially learn a mapping between an input and its response. Whereas, task-oriented dialogue agents typically serve a user's information need by taking a natural language utterance as input and by performing a query in a knowledge base or the web using the predicted intent and finally generating the response from the result~\cite{chen2017survey,radlinski2017theoretical}. In contrast, a robotic dialogue agent needs to understand the semantics of an utterance, by parsing it to a structured and logical form. In the robotics domain, utterances are often short, incomplete and ambiguous that lead to multi-turn dialogues. 

The task-oriented dialogue agents often engage in multi-turn dialogues to extract unspecified arguments~\cite{patidar2018automatic}. However, such end-to-end conversational systems that allow a human to naturally interact with a robot for specifying tasks are rare. Although there are approaches that train a deep neural network to learn end-to-end task-oriented dialogues~\cite{chen2017survey}, it is difficult to collect such training data for robotic task disambiguation and information elicitation. Also, significant efforts are required to adopt such models to work on robots with different manipulation capabilities, deployed in different environments. In the robotics domain, dialogue strategies to elicit missing information has been proposed in~\cite{padmakumar2017integrated,thomason2015learning}. However, the proposed dialogue agents use restrictive dialogue policies that only accept answers that are expected in a context and does not allow the user to change the dialogue flow. In contrast, we present a mixed-initiative dialogue strategy, where the flow of the dialogue can be decided by both the agent and the human.

Instructing a robot through natural language has been widely investigated, but the interactions are often unidirectional and limited to commands~\cite{antunes2016human,bastianelli2016discriminative,lu2017integrating,misra2016tell,tellex2011understanding}. Many proposed works generate execution plans from natural language instruction, by following a parsing-reasoning-planning pipeline~\cite{antunes2016human,lu2017integrating,liu2018generating}, but the role of dialogue in such a pipeline is not well investigated. Following a similar approach for task understanding, we present a dialogue agent that handles prediction failures and incomplete instructions. 

Dialogue agents for task execution by robots are mostly focused on eliciting missing information~\cite{padmakumar2017integrated,thomason2015learning}, knowledge grounding~\cite{she2017interactive,thomason2019improving} and interactive task learning~\cite{cantrell2011learning,chai2018language}. To the best of our knowledge, dialogue to resolve task prediction failures due to ambiguity and novelty in the instruction is not well investigated. Task prediction failures have been tackled by word similarity measures~\cite{lu2017interpreting} and environment-specific data~\cite{misra2015environment}, but the role of dialogue is neglected. Although a similar natural language grounding system for task disambiguation has been proposed in~\cite{pramanick2019enabling}, we present a novel dialogue strategy for this problem that enables learning from past interactions. This strategy leads to better instruction interpretation, using the help from a human and by asking minimal questions that are also easily understood. Our proposed system can also guide the user to interact effectively by providing appropriate responses against predicted user intent.

\section{SYSTEM OVERVIEW}
\label{sec:overview}
In this section, we briefly introduce the components of ``Task Conversational Agent for Robots (\system)'' and its design philosophy. Although \system is primarily a dialogue agent, it includes components that are tightly coupled with a robot's perception, cognition, and actuation sub-systems. \system consists of four building blocks -- (i) a \textit{dialogue state manager (DSM)} that handles the flow of a mixed-initiate dialogue, (ii) a \textit{task interpreter} that identifies tasks, along with their relevant parameters from natural language instructions, (iii) a \textit{knowledge base (KB)} that stores a world model, including a model of the robot and (iv) a \textit{plan generator} that ensures a valid sequence of actions are generated for an identified task. 

\subsection{Dialogue state manager} The role of DSM is to maintain and redirect the dialogue flow to different dialogue strategies that are designed for specific contexts. It includes a high-level intent classifier that captures the intention of the user at every point of interaction. Based on the user's intent and the context, DSM forwards the dialogue to the designated states. The intent classifier and the dialogue strategies are described in Section~\ref{sec:dialogue}.

\subsection{Task interpreter}
\label{sec:task-interpreter}
Tasks are given to the robot as natural language instructions, which are often ambiguous and incomplete. Understanding the meaning of such instruction involves determining the type of the task and the corresponding arguments specified in the instruction. However, there can be numerous kind of utterances that may not be task instruction, e.g., a simple statement, a question, etc. So, \system uses a high-level intent classifier to classify the spoken phrase. 

After an utterance is classified as a task instruction by the high-level intent classifier, the type of the task conveyed by the instruction and the arguments mentioned in it, which are required for the physical execution of the task, are predicted. This understanding is enabled by both the linguistic structure of the instruction and the context inferred from the world model. The task and argument types are modeled using the theory of \textit{frame semantics}~\cite{baker1998berkeley}. It models a task as a \textit{frame} that has an unambiguous goal, and the arguments or the parameters as \textit{frame elements}. We use conditional random field (CRF) models for this sequence prediction. 

For a given utterance $S$ as a sequence of words, $S=\{w_1,w_2,\dots, w_n\}$, we define a model that predicts a label $t_i$ for each $w_i \in S$. We do labeling of task type and the associated parameters in two sequential steps. At the first step, task type is estimated using a probability distribution over the set $T'=T \cup O$.  Here, $T$ is the set of known task types and $O$ contains a single label $o$ for the words that do not express a task. To predict $t_i$, we use both lexical and grammatical features of the word and its contextual words. The features include lemma, parts of speech and syntactic relations. Specifically, the verbs that represent a \textit{frame} are discriminated from the other words using these features. The features are extracted using a general-purpose NLP engine, Spacy\footnote{https://spacy.io/}. The CRF model for task type prediction estimates the following conditional probability distribution.
$$P(t_{1:n}|w_{1:n})= \alpha \exp \bigg \{\sum_{i=0}^n \sum_{j=0}^k \lambda_j f_j(S,i,t_{i-1},t_i)  \bigg \},$$
where  $\alpha$ is a normalization factor, $f_j$ is the $j^{th}$ feature function, $\lambda_j$ is the weight of the $j^{th}$ feature function, and k is the number of such feature functions. During training, the weights of the feature functions are learned using a stochastic gradient descent algorithm. During inference, the maximum likelihood of $t_i$ is used to label the words,
$$T_p=\argmax_{T'} P(T'|S).$$


Similarly, the CRF model for argument extraction estimates the following conditional probability distribution.

\vspace{-0.3cm}
\begin{align*}
P(a_{1:n}|w_{1:n})= \alpha \exp \bigg \{\sum_{i=0}^n \big (\sum_{j=0}^{l-1} \lambda_j f_j(S,i,a_{i-1},a_i)  \\+\lambda_{l}\; g(S,i,T_p) \big ) \bigg \},
\end{align*}
where along with $l-1$ feature functions, the feature function $g$ with the weight $\lambda_l$ is used to associate a task type label with each word. The function $g$ is defined as the following,
\[
g(S,T_p,i) = 
\begin{cases}
\phi,& \text{if } t_i \in T\\
t_j,& \text{else if } t_j \notin T \text{ and } j>i.
\end{cases}
\]
The extracted labels $a_{1:n}$ of the arguments are grounded to known objects using the knowledge base.

\subsection{Knowledge base} 
The knowledge base (KB) stores the model of the environment where the robot is operating and a model of the capabilities of the robot in a formal representation so that reasoning can be performed over the knowledge. The KB has the provision of updating the knowledge with the perceived changes in the environment. The KB also provides the task context that is taken into consideration while generating the planning problem for the planner. This task context is found by updating the KB after actions are performed by the robot by reasoning over the post-conditions of the action sequence with the world model.

\subsection{Plan generator} 
To execute a task, a robot often needs to perform a sequence of sub-tasks that are supported by its manipulation capabilities. A plan is a sequence of such sub-tasks that satisfies the intended goal of the task. For example, to perform a bringing task, the robot has to perform a sequence of movement, picking and placing actions and this sequence is governed by the world state. A task specified in an instruction can be assumed to change a hypothetical state of the world (initial state) to an expected state (goal state). The plan generator uses templates to encode both the initial and goal conditions that are stored as a conjunction of fluents expressed in first-order logic. The planning problem is generated in the PDDL formal language~\cite{mcdermott1998pddl}. First, the appropriate template is selected using the predicted task type and then grounding the variables in the templates using the arguments mentioned in the instruction. The arguments are validated with the current state of the world model given by KB. For the instructions with multiple tasks, the tasks are assumed to be planned and executed in serial order, preserving the context across the tasks. For such instructions with multiple tasks, arguments are often referred by pronouns. For example, in the instruction:\textit{``Take a pen and bring it to me''}, the argument \textit{pen} in the taking task is referred by the pronoun \textit{it} in the next bringing task. We use a co-reference resolver\footnote{https://github.com/huggingface/neuralcoref} to replace such anaphoric references. After generating the planning problem, we use the FF planner~\cite{hoffmann2001ff} to generate the required plan.

\section{DIALOGUE STRATEGY}
\label{sec:dialogue}
\begin{figure}
	\centering
	\includegraphics[height=7.4cm,width=6cm]{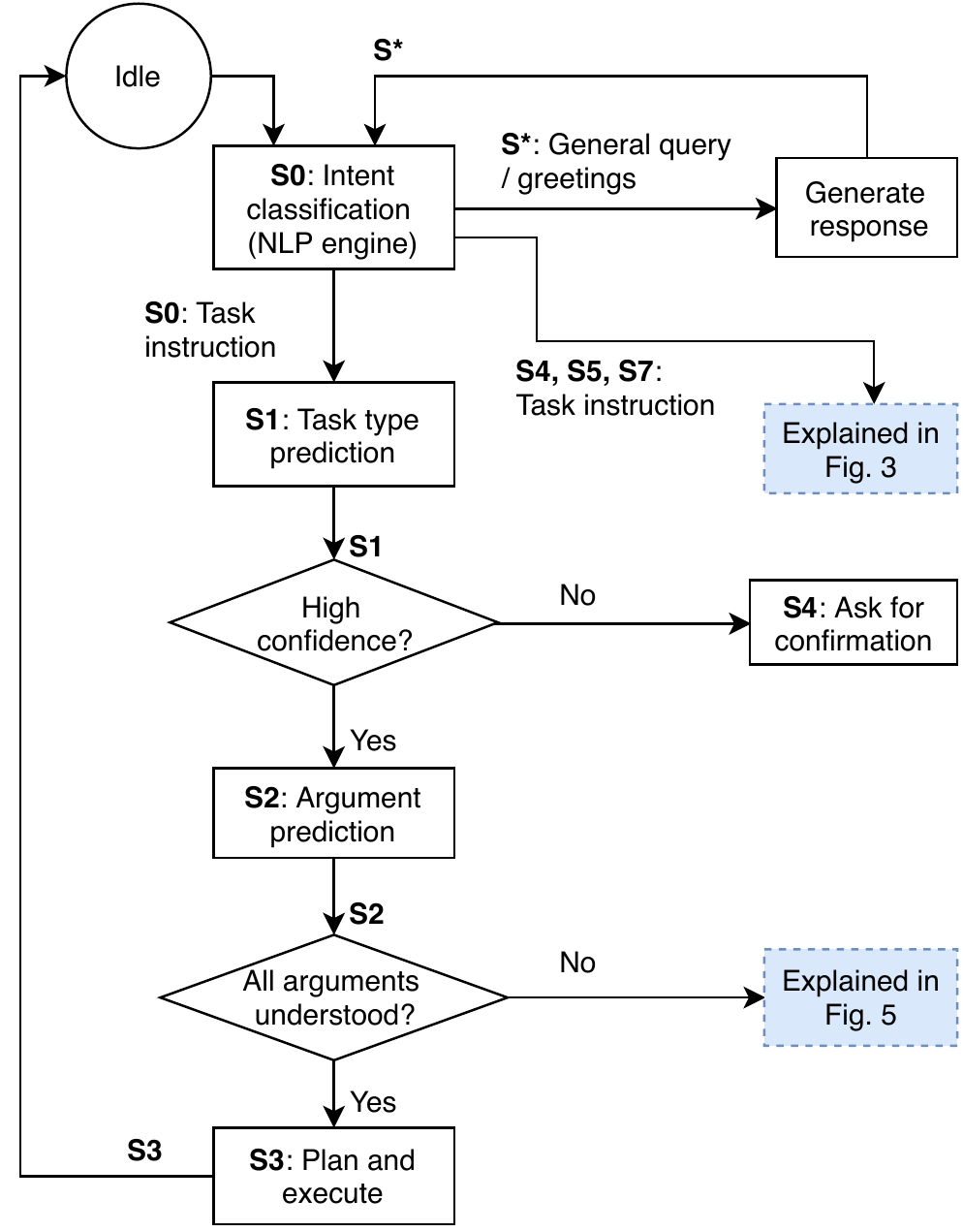}
	\caption{Overview of the dialogue flow in \system.}
	\label{fig:overall-flow}
\end{figure}

The overall flow of dialogue is modeled as a state machine shown in Fig.~\ref{fig:overall-flow}. The state machine consists of several dialogue strategies that are designed to have a concise and meaningful conversation with a human user. We present the strategies as a guideline of \textit{what} needs to be asked in a particular situation and \textit{how} it should be asked. In the following, we present the primary components of this state machine.

\subsection{Intent classification}
\system uses an intent classifier for the probabilistic prediction of a user's intention, given an utterance. We model the intent predication as a text classification problem. The intent classifier takes the training data $D$ as a set of the pairs of an utterance $X_i$ and the corresponding intent $Y_i$, i.e $D=\{x_i,y_i\}_{i=1}^N$. We use a logistic regression classifier, trained with a stochastic gradient descent algorithm and word n-grams as the features for the prediction. During inference, given an utterance $x$, its intent $y$ is found as,
\[y=\argmax_{y_i \in D}  P(y_i|x).\]

The intents recognized by \system are shown in Table~\ref{table:intents}. Initially, the user takes the initiative to start the dialogue, which is shown as the \textit{Intent classification (S0)} state. If the intent is recognized as a task instruction, the system goes ahead to the \textit{Task type prediction (S1)} state. For the intent \textit{question\_own\_location}, the KB is consulted for the robot's current location and a response is generated. For the intent \textit{question\_on\_self}, the manipulation capabilities of the robot are listed. For a \textit{wh\_general} intent, \system responds that it is incapable of answering such questions. For the greeting intents, a response is selected randomly from a set of pre-defined responses.
\begin{table}[t]
	\centering
	\caption{High-level intents recognized by \system.}
	\label{table:intents}
	\begin{tabular}{|l|l|}
		\hline
		\textbf{Intent} & \textbf{Description} \\ \hline
		welcome\_greetings & General greetings of a welcoming note\\ \hline
		question\_on\_self & Questions about the robot's capabilities\\ \hline
		wh\_general & Questions unrelated to the robot\\ \hline
		instruction & Instruction to perform a task\\ \hline
		question\_own\_location & Questions about the robot's current location\\ \hline
		bye\_greetings & Statements denoting the user wants to leave\\
		\hline
	\end{tabular}
\end{table}

If a task is predicted with high confidence, then \system goes ahead to the \textit{Argument prediction (S2)} state, otherwise \system takes the initiative to start a dialogue and the dialogue strategy for the same is described in Section~\ref{subsec:Task-disambiguation-dialouge}. If the extracted arguments are valid and fulfill the requirement for the planing problem generation template, it goes to the \textit{Plan and execute (S3)} state. Otherwise, \system engages in a dialogue to elicit the missing information, as explained in Section~\ref{subsec:Argument-elicitation}.

The user can change the goal of the dialogue at any state by expressing his/her intention to do so. For example, when \system is asking to confirm its task type prediction, the human can give a new task or modify the arguments instead of giving an answer. The change in the initiative is determined by the intent classifier and the corresponding state transition in the state machine. 

\subsection{Task disambiguation by dialogue}
\label{subsec:Task-disambiguation-dialouge}
\begin{figure}[t]
	\centering
	\includegraphics[height=5.2cm,keepaspectratio]{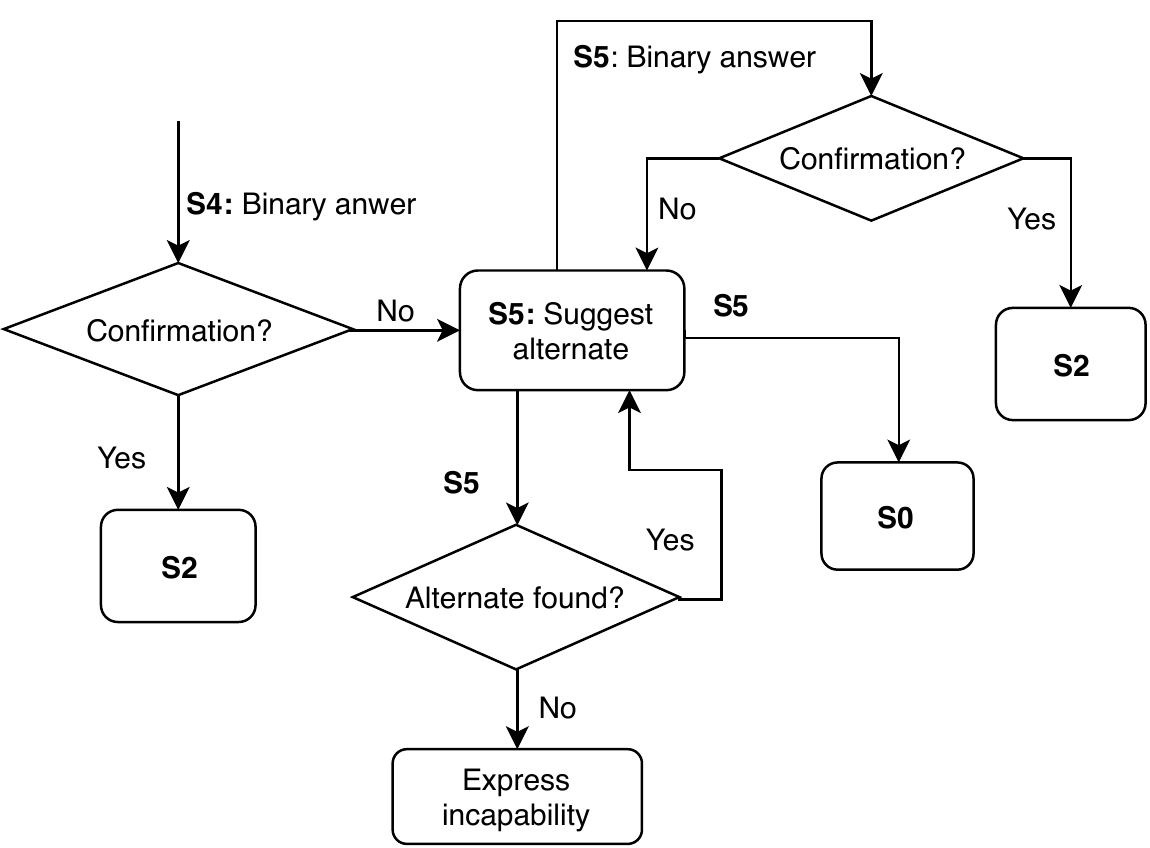}
	\caption{Dialogue strategy to resolve task prediction failure.}
	\label{fig:task-amiguity}
\end{figure}

The task type prediction model is a probabilistic classifier that is subject to uncertainty. Typically, task identification models are trained with the features around the verb present in the instruction and with a limited set of such training examples~\cite{bastianelli2016discriminative,misra2016tell,pramanick2019enabling}. During prediction, the models can encounter novel verbs and ambiguous sentence structures that may lead to mispredictions. Also, the features are extracted using probabilistic classifiers and their uncertainties are propagated to the task prediction model. It is natural to confirm such predictions from the human if the confidence of the prediction is low~\cite{thomason2015learning}. \system asks the user to validate a low confidence prediction before forwarding to the argument prediction state.  We use the likelihood of the task type estimated by the task prediction model as the confidence. If the prediction is confirmed by the user then \system proceeds to the argument prediction state. If the prediction is stated to be incorrect, then the intended task type is determined by engaging in a dialogue. Also, it has to be determined whether the robot is actually capable of performing the task, as the human may not be aware of the same. Fig.~\ref{fig:task-amiguity} depicts this dialogue strategy.

However, directly asking the user to specify the task type, as proposed in~\cite{thomason2015learning} is impractical, because the user may not know the task types known to the robot. It is also difficult for a novice user to infer the convention that is used to define the task types. Generally, in such situations, it is better to provide the user with specific choices~\cite{radlinski2017theoretical}. We use a dialogue strategy to ask the user about alternate task types, also making sure that the questions are easily understood. The strategy asks the user about the similarity of the given instruction with the known set of tasks. The user is able to give a binary yes/no answer to these questions so that the answer also becomes unambiguous for \system.

In a practical scenario, a robot needs to understand tens of task types. If the dialogue suggests them one by one, the human experience will degrade badly. So the robot needs to suggest alternative task types in the order of their likelihood of being the true task type. We propose a method to estimate the likelihoods by exploiting the training data given to the robot, which can also include the conversation history experienced by the robot. Specifically, we hypothesize that if the task type can not be determined from the features, the probable arguments present in the instruction can provide evidence for the task type. In this case, the argument types present in the instruction are predicted without considering the task type associated with the words. Specifically, given a sentence $S$, we estimate a conditional probability distribution over the set of task types $T$, i.e., $P(T|S)$. Then the task types are ranked using their probabilities and the dialogue strategy asks the questions using the ranked list. After asking about all the task types in the list, \system determines that the robot is unable to perform the task. If the number of task types is very large, then a probabilistic threshold can also be used to express the incapability earlier. The model for predicting the argument types is also realized as a CRF that estimates the following.
$$
P(a'_{1:n}|w_{1:n})= \alpha \exp \bigg \{\sum_{i=0}^n \sum_{j=0}^{l} \lambda_j f_j(S,i,a'_{i-1},a'_i) \bigg \},
$$
where $a'_i$ is the predicted label of an argument type, for the word $w_i$. This model uses the same features as the argument extraction model, except the task type association feature function.

The predicted labels $a'_{1:n}$ are used to determine the set of argument types $A'$ present in $S$. Given the training data $D$, as $m$ instances of annotated instructions $D=\{I_k\}_{k=0}^m$, we extract the set $A^D_k$ for each instance $I_k$. Then the number times a task type $t\in T$ satisfies the condition $A' \subset A^D_k$, is counted for all the $m$ annotation instances. The counts are normalized using a softmax function to estimate the probability distribution $P(T|S)$. To enable learning from past interactions, $D$ also includes the annotated history of the instructions successfully planned by \system. Furthermore, during the normalization, the counts from past interactions can be given more weight to give preference to user-specific vocabulary over the offline training data. 

While asking about the task type prediction and the alternatives, \system needs to convey the meaning of the task type to the user through the question. The question needs to be carefully crafted, so that a user who is not aware of the terminologies used by the robot, can understand the question.
As an example, consider the ambiguous instruction: \textit{``Put on the display''}. This instruction is predicted with low confidence as a task of changing the state of a device because of the ambiguous verb \textit{Put}, but it could also mean a placing task. However, a question like \textit{``Do you want me to do a state change task?''} is less likely to be understood properly. Instead, we use templates to frame the questions that preserve the similarity of the question with the original instruction. Examples of the templates are shown in Table~\ref{table:task-templates}.

The underlined words shown in the table denote unfilled argument slots. The slots are filled by extracting the arguments from the instruction using the task type for which the confirmation is being asked. For the same example, \system frames the question \textit{``Do you want me to turn \textbf{on} the \textbf{display}?''}, which is better understood. If a slot is unfilled, i.e., not mentioned in the instruction, a generic phrase denoting the argument type is used to fill the argument slot. For example, to ask if the instruction conveys a placing task, the question is framed as \textit{``Do you want me to put the \textbf{display} in \textit{somewhere}?''}.
\begin{table}
	\centering
	\caption{Question templates for task disambiguation.}
	\begin{tabular}{|l|l|}
		\hline
		\textbf{Task type} & \textbf{Template} \\ \hline
		Motion & Should I move to \textit{\underline{location}}? \\ \hline
		Taking & Do you want me to pick up \textit{\underline{object}}? \\ \hline
		Bringing & Should I bring \textit{\underline{object}} to \textit{\underline{location}}? \\ \hline
		Change-state & Do you want me to turn \textit{\underline{intended-state}} the \textit{\underline{device}}?\\ \hline
		Placing & Do you want me to put the \textit{\underline{object}} in \textit{\underline{location}}? \\ \hline
	\end{tabular}
	\label{table:task-templates}
\end{table}{}

\subsection{Argument elicitation}
\label{subsec:Argument-elicitation}
\begin{figure}
	\centering
	\includegraphics[width=0.7\linewidth,height=6.5cm]{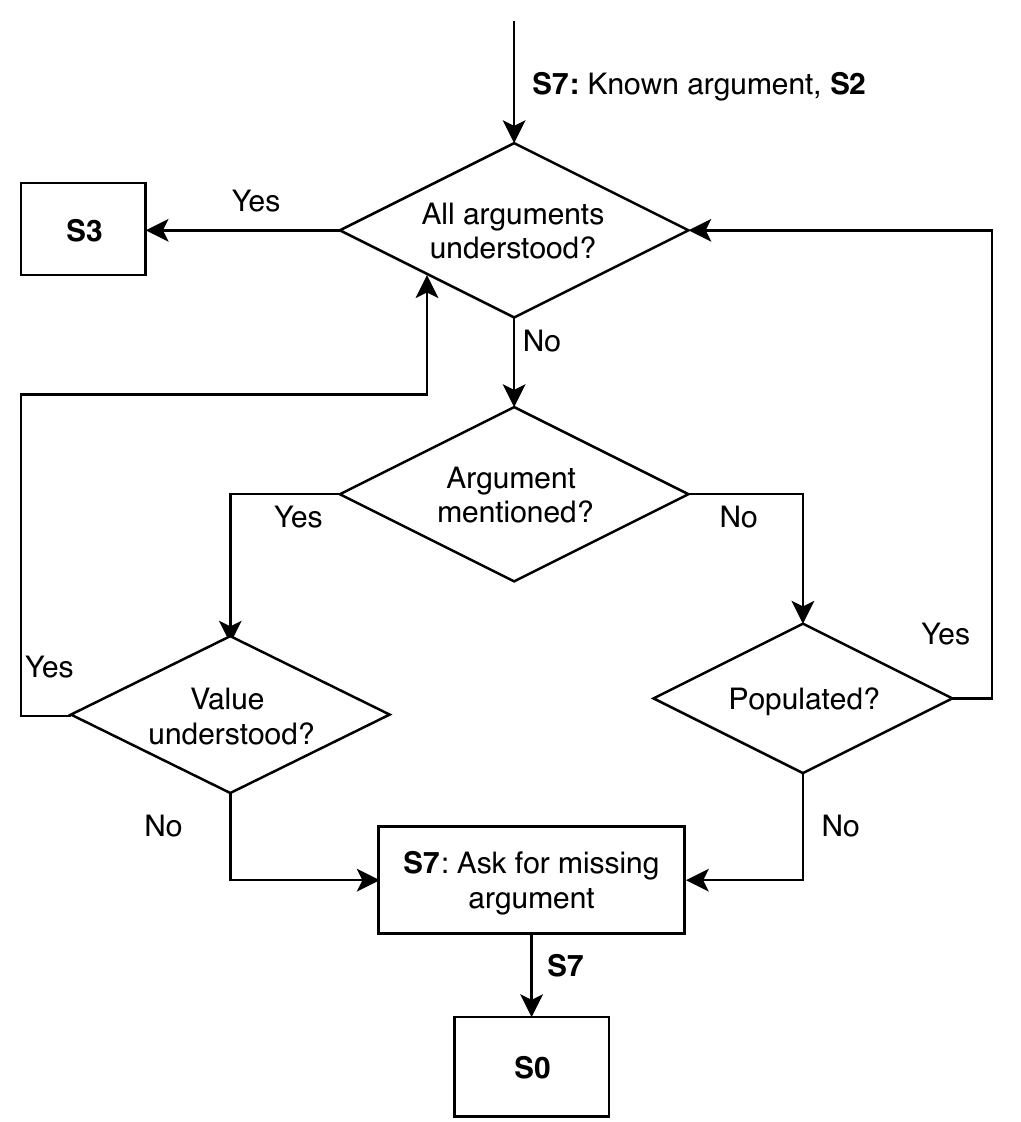}
	\caption{Dialogue to validate arguments before planning. }
	\label{fig:argument-resolution}
\end{figure}

Before generating the planning problem, TCAR validates the required arguments for the task. This list of arguments depends upon the task template and the planning context given by KB. For example, if the robot is instructed to bring an object to another location, the source location of the object needs to be specified if that information is neither present in the instruction, nor stored in KB. But if the robot is already holding the object, for the same instruction, the source location need not be mentioned. It may also happen that the argument itself can be ambiguous. For example, if there are multiple doors in the room, for an instruction to go to a door, the robot asks for disambiguation by showing the choices. Fig.~\ref{fig:argument-resolution} shows the dialogue strategy for eliciting the argument information. 

For all the arguments required for the given task, \system checks whether they are mentioned in the instruction using the argument prediction model. If an argument is mentioned, i.e., the type of argument is known but the value is not stored in KB, \system asks to provide a valid value for the argument. Otherwise, \system checks if the argument can be populated using the world model from KB. If not so, \system asks the user to specify the missing information. Again, the questions are generated using templates and some of the question templates are shown in Table~\ref{table:argument-templates}. For an argument that is shared across multiple task types, a generic template is used that uses an appropriate synonym of the task to generate the question. For the unique arguments (used only in a certain task type), we use predefined questions.
\begin{table}
	\centering
	\caption{Question templates to elicit missing arguments.}
	\begin{tabular}{|p{2cm}|p{2.2cm}|p{3cm}|}
		\hline
		\textbf{Task type} & \textbf{Missing argument} & \textbf{Template} \\ \hline
		Taking, Bringing & Source location & From where do I \textit{\textbf{Verb}} it? \\ \hline
		Bringing, Placing & Goal location & Where should I \textit{\textbf{Verb}} it? \\ \hline
		Change-state & Device & Which device do I turn on/off? \\ \hline
		Searching & Area to search & Where do I search for it? \\ \hline
	\end{tabular}
	\label{table:argument-templates}
\end{table}

\subsection{Dialogue Session continuation}
\begin{figure}
	\centering
	\includegraphics[width=0.8\linewidth,keepaspectratio]{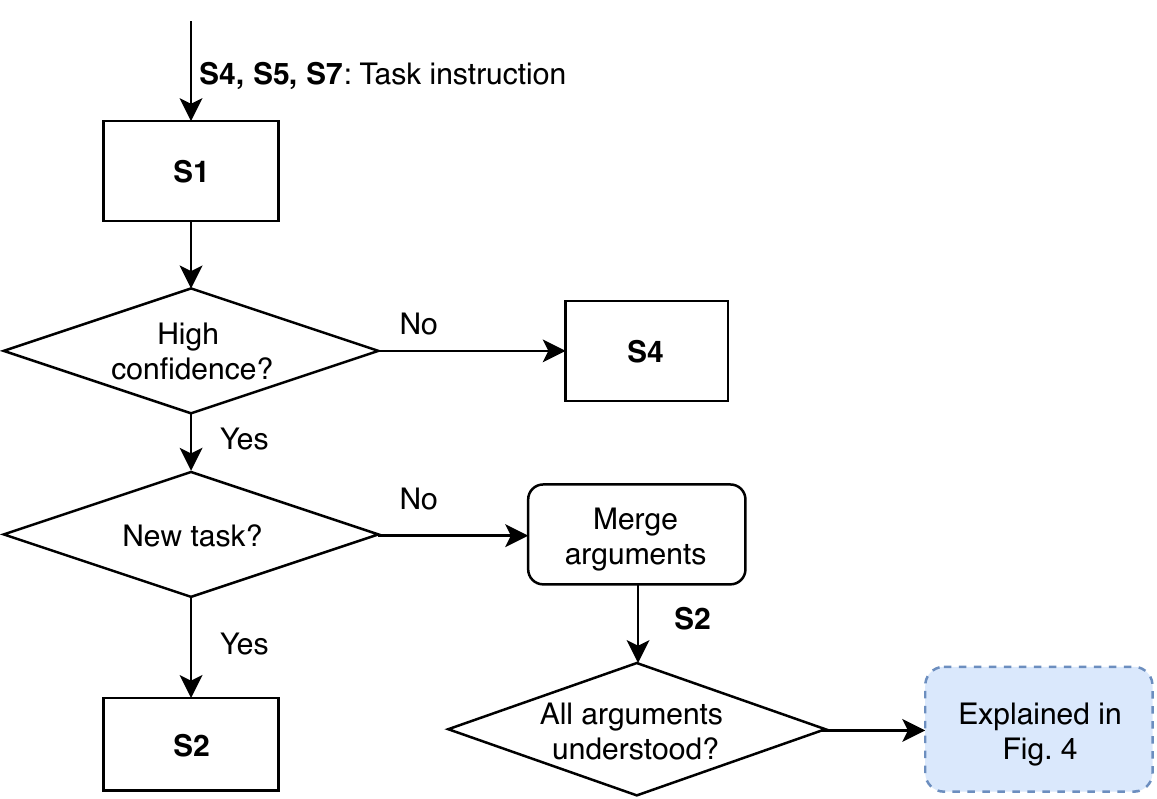}
	\caption{Ensuring dialogue continuity using the context set by the previous task.}
	\label{fig:task-context}
\end{figure}

We have equipped \system with the capability to maintain dialogue continuity even if unexpected answers are given. \system expects binary answers while asking to confirm a task type prediction in the states \textit{S4} and \textit{S5} shown in Fig.~\ref{fig:overall-flow} and Fig.~\ref{fig:task-amiguity}, respectively. But instead of a binary answer, the user may rephrase the instruction, possibly with pronoun references of the arguments. As an example, for the question \textit{``Do you want me to turn on the display?''}, the user may give the answer as a task: \textit{``Turn it on''}, also referring to the noun \textit{display} by the pronoun \textit{it}. Similarly, when \system asks to provide a missing argument in the state \textit{S7} (Fig.~\ref{fig:argument-resolution}), instead of answering in a word or phrase, the answer may re-iterate the original instruction with the required argument. For example, in response to the question \textit{``From where do I take it?''}, the answer can be given as \textit{``take it from table''}. Moreover, the user may give a new task or may simply intend to end the conversation. Our dialogue strategy can also resolve such unexpected answers using the notion of session continuation. We define a session as a unit of the conversation that starts from an utterance given by the human and ends with either a task execution, failure to understand an utterance or a \textit{bye\_greetings} intent. In the same session, the human is expected to talk about a single task, but the session can be preempted by providing a new task. The dialogue strategy is shown in Fig.~\ref{fig:task-context}. 

We use the same intent classifier as discussed earlier to determine the intent conveyed by the answer provided in the states \textit{S4, S5}, and \textit{S7}. If the intent is classified to be an instruction, then the task type is predicted. If the task type is predicted with high confidence and is of the same type as the one in the current session, \system goes to the argument validation state, adding the new arguments (if any) to the task and continues the strategy described in Section~\ref{subsec:Argument-elicitation}. For a low confidence prediction, the dialogue is continued using the strategy described in Section~\ref{subsec:Task-disambiguation-dialouge}. While merging the arguments, we use a co-reference resolver to replace the pronoun references with the arguments mentioned in the original instruction.

\section{EVALUATION}
\label{sec:eval}
In this section, we present the results of an automated quantitative evaluation of \system using a dataset and a subjective evaluation by human users.

\subsection{Quantitative evaluation}
We train the CRF models for instruction understanding using the HuRIc dataset presented in~\cite{bastianelli2014huric}. From the dataset, we sample 481 instructions annotated with the tasks and the mentioned arguments after removing the task types that have very few (less than 5) samples. This results in a total of 9 task types and 11 argument types. We split the dataset into 75\% training set and 25\% test set and report the accuracy of the models for the test set in Table~\ref{table:result-huric}. Though the task type prediction and the argument extraction models performs well, the accuracy of argument type prediction, without using the task type association feature is not high. However, the moderate inaccuracy of this model does not hinder the end result as the dialogue strategy for task disambiguation does not use the output of this model directly.
\begin{table}
	\centering
	\caption{Classification report of the CRF models for the HuRIc dataset.}
	\label{table:result-huric}
	\begin{tabular}{|p{3.4cm}|p{1.1cm}|p{1.1cm}|p{1.1cm}|}
		\hline
		\textbf{CRF model} & \textbf{Precision} & \textbf{Recall} & \textbf{F1 Score} \\ \hline
		Task type prediction & 0.93 & 0.90 & 0.91 \\ \hline
		Argument extraction & 0.93 & 0.92 & 0.92 \\ \hline
		Argument prediction without task type information & 0.75 & 0.75 & 0.72 \\ \hline
	\end{tabular}
\end{table}

We have evaluated the pipeline of plan generation from instructions using the Rockin@Home\footnote{http://rockinrobotchallenge.eu/home.php} dataset that has been collected from several competitions for assessing instruction understanding capabilities by a domestic service robot. In our evaluation, task planning is successful when the task type and arguments are correctly predicted and the generated planning problem results in a valid plan by the FF planner~\cite{hoffmann2001ff}. We compare our system against two baselines. The \textit{Baseline-ND} system does not use any dialogue to interpret a task. The \textit{Baseline-AD} system uses dialogue for argument elicitation, but only for the arguments not mentioned in the instruction. During the evaluation, \system uses argument elicitation dialogue with the provision of populating missing arguments using the KB along with co-reference resolution. The human responses are automated by a simulation that provides the correct missing argument only if it is required and can't be inferred. The responses provided in this simulation are always direct answers, in a word or phrase. The same task identification and argument extraction models are used in \system and the baselines. 
\begin{table}[t]
	\centering
	\caption{Plan generation results for the Rockin@Home dataset.}
	\label{table:task-identification-accuracy}
	\begin{tabular}{|l|l|}   
		\hline
		\textbf{System} & \textbf{Plan generated} \\
		\hline
		No dialogue (\textit{Baseline-ND}) & 183 (42.5\%) \\ \hline
		Naive argument elicitation (\textit{Baseline-AD}) & 334 (77.5\%)\\ \hline
		\system & 392 (90.9\%)  \\
		\hline
	\end{tabular}
\end{table}

We report the plan generation accuracy for the dataset in Table~\ref{table:task-identification-accuracy}. The task identification model is able to identify 420 (95.6\%) out of the 431 tasks specified in 385 instructions, containing 1.12 tasks per instruction. The \textit{Baseline-ND} system is able to generate a valid plan for only 42.5\% tasks because many of the instructions were incomplete. The \textit{Baseline-AD} generates plans for 75.4\% of the tasks, outperforming \textit{Baseline-AD} by a large margin, but it fails for instructions with multiple dependent tasks that requires inferring arguments from the task context. \system generates 90.9\% of the tasks that match closely with the accuracy of the task identification model. For some of the correctly identified tasks, plan generation fails due to argument parsing failures. We can not evaluate the task disambiguation dialogue strategy for the incorrectly identified tasks because of the similarities in the task types between the Rockin@Home and the HuRIc dataset. Instead, we present the results of a user study to evaluate this in the following sub-section.

\subsection{Subjective evaluation}
We conduct a study with human participants to evaluate \system in a telepresence meeting scenario, where a robot acts as an avatar of the attendee. The goal of the study is to infer how people would interact with \system given that dialogue systems are generally perceived as question-answering agents and its application for instructing robots is not well known to the public. We also hypothesize that there is a high expectation of interaction quality from conversational systems because of the popular usage of voice-based personal assistants and people would expect similar responding capabilities from \system even though its applicability is very different. The second goal is to assess \system's language understanding capability for novel utterances and see whether the dialogue strategies we described in Section~\ref{sec:dialogue} can guide the participants to successful task executions.

For the experimentation, we develope a graphical interface that allows a participant to type in the utterances along with a window showing the interaction and another window showing animations in a simple simulated environment as the robot executes a task. From our experiments, 12 participants (5 female, 7 male) with ages in between 25-48\textit{ (mean(m)=32, standard deviation(sd)=8.1)} volunteered for the study. All the volunteers have a bachelors degree and higher education except one person who has a high-school education. None of them is a native English speaker, but well conversant in English. No volunteers have any prior experience of working in robotics or natural language processing. On a scale of 1 to 10, the average knowledge of how a robot works is about 4.91 (based on their rating).

To validate our hypothesis, we neither explicitly reveal \system's language understanding nor manipulation capabilities. Instead, the participants are stated the following: \textit{``A mobile telepresence robot can attend a meeting on your behalf. Interact with the robot using the chat window, imagining you are remotely using it.''}. The participants are asked to interact with the system for a maximum of 10 minutes. We partition the total interaction of each user into sessions. A session starts with a greeting from \system, and ends when any of the following conditions are met -- (i) a task is executed, (ii) \system predicts a \textit{bye\_greetings} intent,  (iii) \system can not understand the intent of the utterance, (iv) \system expresses that it is incapable of performing the task. We record a total of 126 sessions for 12 participants (\textit{m=10.5, sd=6.1}). In total, the participants use 261 utterances (\textit{m=21.8, sd=11.5}) with an average of 2.07 utterances per session (\textit{sd=1.78}). Out of the 12 participants, 7 of them ask unrelated questions (having a \textit{wh\_general} intent) in their first sessions. The participants ask a total of 12 questions (38.7\%) out of the 31 first-session utterances. This supports our hypothesis that initially \system is being perceived as a question-answering dialogue agent.

The task understanding model of \system is subjectively evaluated by recording the number of tasks provided by participants and the corresponding number of successful plan executions. Out of 113 given tasks, a total of 85 plans are generated and executed in simulation, which results in an accuracy of 75.2\%. We also notice that for 38 out of the 85 successful tasks (44.7\%), \system needs to ask further questions to elicit missing information and to perform task disambiguation. This evidence indicates that natural language instructions are often incomplete and ambiguous, which require further questions to be fully understood. 

For the task disambiguation dialogues of successful executions, \system asks an average of 2.3 questions (\textit{sd=1.06}) and for the tasks beyond its capability, \system has to ask about all the five tasks. We also measure the amount of time a participant spent to give an answer after a question is asked by \system. We find that for the cases of successful task execution of the first task instruction, participants spend an average of 38.3 seconds (\textit{sd=33.1}). For subsequent tasks by the same participant, they spend 13.6 seconds (\textit{sd=29.05}) on average per questions. This is an indication that the participants learn to interact more effectively from their dialogue experiences with \system.  

\begin{figure}[t]
	\centering
	\includegraphics[width=\linewidth]{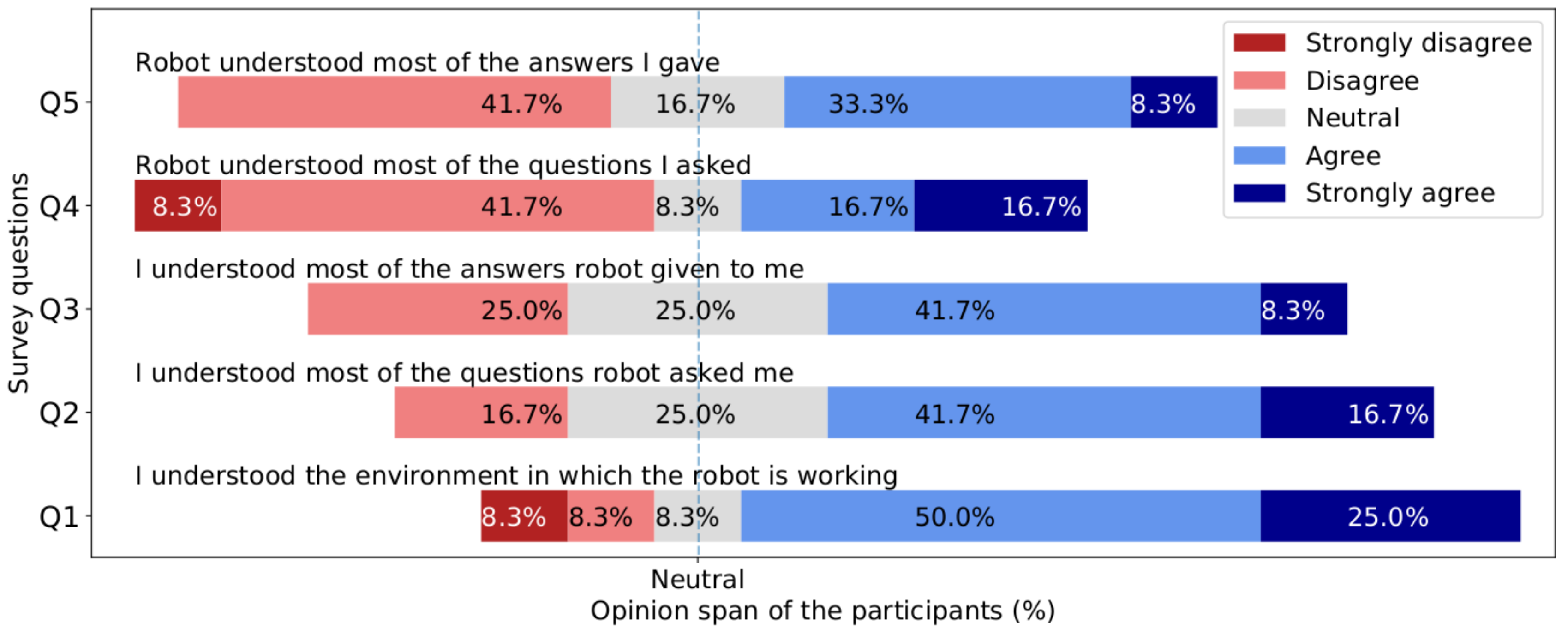}
	\caption{User experience on different aspects of the interaction.}
	\label{fig:likert}
\end{figure}

We also ask the participants to fill up a questionnaire about their experience with \system and to provide suggestion to improve the interaction quality. The participants are asked to rate various aspects of the dialogue using a \textit{Likert scale}. The aspects and the recorded ratings are shown in Fig.~\ref{fig:likert}.
The ratings show that the participants mostly understand the environment (\textit{Q1}). Most of the participants understand the questions \system asked (\textit{Q2)} and also the answers \system has provided (\textit{Q3}). Many of them have felt that \system can not understand many of their questions (\textit{Q4}). One reason for this is the participants are not aware of the agent's capabilities and they perceive it as a question-answering agent, leading to many irrelevant questions that are not properly captured by the list of intents recognized by \system. Also, many participants have suggested that \system should list down its capabilities before starting the dialogue. Even so, the results indicate that understanding user utterances is the most important and also a very challenging part for the development of a robotic conversational system. Some of the participants have felt that \system understands the answers they have given (\textit{Q5}), while a similar percentage of participants have felt otherwise.

\section{CONCLUSIONS}
\label{sec:con}
Providing task instructions to a cohabitant robot through natural conversation adds to the usability and acceptability of the robot, especially for a non-expert user. We present a conversational agent for robots that understands tasks that are specified in natural language. The agent is also capable of guiding a novice user to specify tasks more effectively through a meaningful conversation. We propose several dialogue strategies employed in the conversational agent to understand novel or ambiguous instructions and to seek help by asking minimal questions. In the future, we would like to include gesture interpretation for multi-modal instruction understanding and also evaluate our system with people from diverse cultural and educational backgrounds.

\balance
\bibliographystyle{IEEEtran}
\bibliography{main}

\end{document}